\pdfoutput=1
\documentclass[11pt]{article}

\usepackage[review]{acl}
\usepackage{caption}
\usepackage{subcaption}
\usepackage{amsmath}
\usepackage{bm} 
\usepackage{url}
\usepackage{float}
\usepackage{times}
\usepackage{amssymb}
\usepackage{latexsym}
\usepackage{todonotes}
\usepackage[T1]{fontenc}
\usepackage[utf8]{inputenc}

\usepackage{microtype}
\newcommand{\tildea}[1]{\overset{\sim}{#1}}
\newcommand{\tildeb}[1]{\stackrel{\sim}{\smash{#1}\rule{0pt}{1.1ex}}}
\title{RubCSG at SemEval-2022 Task 5: Ensemble learning for identifying misogynous MEMEs}

\author{Wentao Yu, Benedikt Boenninghoff, Jonas Roehrig, Dorothea Kolossa \\
 Institute of Communication Acoustics, Ruhr University Bochum, Germany \\
  \texttt{\{wentao.yu, benedikt.boenninghoff,} \\
  \texttt{jonas.roehrig, dorothea.kolossa\}@rub.de}}

\begin{document}
\maketitle
\begin{abstract}
%Hate speech against women is rampant for a long time. Identifying and blocking hate speech seems to be the most effective way for the time being. We propose an ensemble system based on various single-modal and multi-modal model architectures for the SemEval-2022 Task 5: MAMI-Multimedia Automatic Misogyny Identification, which provides a dataset to develop and optimize the detection algorithms for English hateful memes against women. This competition consists of two sub-tasks: sub-task A is for misogynous meme identification. Sub-task B aims to classify the misogynous meme into potential overlapping categories: stereotype, shaming, objectification, and violence. The annotation was manually annotated by crowdsourcing platforms.
%Our single-modal models consider the transformer model and graph convolutional network structure. While the multi-modal models use our proposed model fusion network. %on the official test set for both binary classification task (sub-task A) and multi-label classification task (sub-task B). 
%Finally, we got 0.755 macro-average F1-score (11th) in binary classification task (sub-task A) and 0.709 weighted-average F1-score (10th) in multi-label classification task (sub-task B).
 
This work presents an ensemble system based on various uni-modal and bi-modal model architectures developed for the SemEval 2022 Task 5: MAMI-Multimedia Automatic Misogyny Identification.
The challenge organizers provide an English meme dataset to develop and train systems for identifying and classifying misogynous memes. More precisely, the competition is separated into two sub-tasks: sub-task A asks for a binary decision as to whether a meme expresses misogyny, while sub-task B is to classify misogynous memes into the potentially overlapping sub-categories of stereotype, shaming, objectification, and violence. 
For our submission, we implement a new model fusion network and employ an ensemble learning approach for better performance. With this structure, we achieve a 0.755 macro-average F1-score (11th) in sub-task A and a 0.709 weighted-average F1-score (10th) in sub-task B.\footnote{Code available at: \url{https://github.com/rub-ksv/SemEval-Task5-MAMI}.}

\end{abstract}

%\begin{keywords}
%  Ensemble \sep
%  BERT \sep
%  multimodal \sep
%  misogynous memes
%\end{keywords}
\section{Introduction}\label{intro}
 Hate speech against women remains rampant despite many efforts at education, prevention and blocking. Misogyny takes place online and offline. Especially on social media platforms, misogyny appears in different forms and has serious implications~\cite{chetty2018hate}. Currently, automated  detection and filtering seem to be the most effective way to prevent hate speech online. However, over the past few years, the rising popularity of memes brought misogyny to a new multi-modal form, which may be more likely to go viral due to their often surprising combinations of text and image that may strike viewers as funny and hence, as eminently shareable. 
 
 The multi-modality of memes also makes automatic detection more challenging. Some memes express their hatred implicitly or through juxtaposition, so they may even appear harmless when considering the text or the image in isolation. SemEval-5 2022 Multimedia Automatic Misogyny Identification (MAMI)~\cite{task5} aims to identify and classify English misogynous memes.
 
 In recent years, the Transformer model \cite{vaswani2017attention} has been widely used in natural language processing (NLP) and image processing. Transfer learning~\cite{torrey2010transfer} with a pre-trained Transformer model can save training resources and increase efficiency with  less training data~\cite{wang2020heck}.
 
 Therefore, in this work, we consider transfer learning to customize three uni-modal models based on the Transformer model: \romannumeral 1) fine-tuning a pre-trained RoBERTa model for classification (BERTC) \cite{liu2019roberta}; \romannumeral 2) training a graph convolutional attention network (GCAN) using the pre-trained RoBERTa model for word embedding; \romannumeral 3) fine-tuning a pre-trained image model, the Vision Transformer (ViT) \cite{dosovitskiy2020image}. Based on these three uni-modal models, four bi-modal models are trained through our proposed model fusion network, namely BERTC-ViT, GCAN-ViT, BERTC-GCAN, and BERTC-GCAN-ViT. All models are evaluated with 10-fold cross-validation. The macro-average and weighted-average F1-scores are employed as the metrics for the sub-tasks. Ultimately, the ensemble strategy is applied on both the dataset- and the model-level (detailed in Section~\ref{ensemble}) for better performance.
 
 The remainder of the paper is structured as follows: Section~\ref{background} introduces the MAMI challenge and related solutions to the task. Our ensemble model is described in Section~\ref{Overview}, followed by the experimental setup in Section~\ref{setup}. Finally, our results are shown and conclusions are drawn in Sections~\ref{results} and~\ref{conclusion}.
\section{Background}\label{background}
The MAMI dataset contains 10,000 memes as the training and 1,000 memes as the test set; all of these are given together with the text transcription as obtained through optical character recognition (OCR). The reference labels are obtained by manual annotation via a crowdsourcing platform. 

The challenge is composed of two sub-tasks: Sub-task A represents a binary classification task and focuses on the identification of misogynous memes, so each meme should be classified as not misogynous (noMis) or misogynous (Mis). Sub-task B, in contrast, presents a multi-label classification task, where the misogynous memes should be grouped further, into four potentially overlapping categories. The dataset class distribution is illustrated in Table~\ref{tab:datadis}.
\vspace{-0.2cm}
\renewcommand{\arraystretch}{1.15}
\begin{table}[!htb]
\caption{MAMI-22 dataset class distribution. \textbf{Mis}: misogynous; \textbf{Shm}: shaming; \textbf{Ste}: stereotype; \textbf{Obj}: objectification; \textbf{Vio}: violence.}
\setlength{\tabcolsep}{5.5pt} % Default value: 6pt
\begin{tabular}{|c | c c c c c|}
\hline
\textbf{Sets}& \textbf{Mis}  & \textbf{Shm}  & \textbf{Ste}  & \textbf{Obj}  & \textbf{Vio} \\ \hline\hline
training set & 5000 & 1274 & 2810 & 2202 & 953 \\ 
test set & 500 & 146 & 350 & 348 & 153 \\ \hline
\end{tabular}
\label{tab:datadis}
\end{table}

%Machine learning for hateful memes identification has been researched for a short time. 
%The multi-modality of the provided memes requires the model to integrate text and image information. %for better performance.
Since the provided dataset contains two modalities (namely, images and texts), an automated approach requires integrating the information from the images with the textual information. However, the OCR-based transcriptions are quite error prone, while the images are often hard to recognize for automatic systems, due, among other reasons, to overlaid text and to the popularity of further changes, such as the composition of multiple sub-images. Consequently, it is challenging to identify the pertinent information of the respective modalities, in order to merge it into a joint classification decision.

%In related work, 
Some researchers have already worked on meme datasets. For example,  \cite{sabat2019hate} created a hateful memes database, using the BERT model to extract a contextual text representation and the VGG-16 convolutional neural network~\cite{simonyan2014very} for image features. Then, text and image representations are concatenated to obtain a multi-modal representation. Facebook also organized a challenge for the identification of hateful memes in 2020~\cite{kiela2020hateful}. The winner of this challenge adopted an ensemble system with four different visual-linguistic transformer architectures~\cite{zhu2020enhance}.
%Important for our work are the Transformer Architecture and Graph Convolutional networks.

The Transformer model has shown excellent performance in many tasks, and it also shows promising results in the above studies, based on its use of the attention mechanism to extract the contextual information within a text. However, its ability to capture global information about the vocabulary of a language remains limited~\cite{lu2020vgcn}, and we hypothesize that this is even more of an issue in the task at hand, due to the very short texts in the given challenge. 

For this reason, we combine a Transformer model with a graph convolutional network (GCN)~\cite{yao2019graph}, which may help to address this issue. GCNs can be understood as a generalization of CNNs, where the data has graph structure and locality is defined by the connectivity of the graph. As input, a GCN receives features that connect to a set of nodes. From layer to layer, the features of a node are updated as weighted combinations of its neighbors\textquotesingle \  features. In our case, the graph is defined as follows: There is a node for every word in the vocabulary and for every document. The collection of nodes is $V = \left \{ D_1, D_2\cdots D_{n_D},W_1, W_2, \cdots W_{n_w}\right \}$, where $D_i$ and $W_i$ indicate the document and word nodes, respectively. $n_D$ is the number of documents and $n_W$ is the number of unique words in the corpus. The edges between word nodes are weighted with the word co-occurrence, the edges between document-word pairs are weighted with the term frequency-inverse document frequency (TF-IDF). 

A fixed-size sliding window with step size 1 is used to gather the %is used on all documents to gather this
word co-occurrence information through the entire dataset. The point-wise mutual information (PMI) is employed to measure the relationship between the words \textit{i} and \textit{j} as follows:
\begin{equation} \label{PMI}
\renewcommand\arraystretch{1.5}
\begin{matrix}
\textrm{PMI} (i, j) = \textrm{log}\frac{p(i, j)}{p(i)p(j)},\\ 
p(i, j) = \frac{N(i, j)}{N},\\ 
p(i) = \frac{N(i)}{N},
\end{matrix}
%\small
%\textrm{PMI} (i, j) = \textrm{log}\frac{p(i, j)}{p(i)p(j)},
%\end{equation}
%\begin{equation} \label{pij}
%\small
%p(i, j) = \frac{N(i, j)}{N},
%\end{equation}   
%\begin{equation} \label{pi}
%\small
%p(i) = \frac{N(i)}{N},
\end{equation}
where $N(i)$ counts the sliding windows in the training set that contain word $i$, $N(i, j)$ is the number of sliding windows that carry both words $i$ and $j$, and $N$ is the total number of sliding windows in the corpus. %Assuming there is a small corpus, which is represented as a set of token ids [0 1 2 5 4 5 6 3 4 6 10]. The window size is 4. So, the corpus could be segmented by the sliding window as follow:
%\begin{equation*} \label{splitwindow}
%\begin{matrix}
%0 &1 &2 &5 \\
%1 &2 &5 &\textbf{4} \\
%2 &5 &\textbf{4} &5 \\
%5 &\textbf{4} &5 &6 \\
%\underline{\textbf{4}} &5 &6 &\underline{3} \\
%5 &6 &\underline{3} &\underline{\textbf{4}} \\
%6 &\underline{3} &\underline{\textbf{4}} &6 \\
%\underline{3} &\underline{\textbf{4}} &6 &10.
%\end{matrix}
%\end{equation*}
%There are eight windows; token id \textbf{4} appears seven times; the word pair (3, 4) appears four times in this corpus (we do not consider the graph direction). Therefore,   $N=8$, $N(4)=7$, $N(3)=4$, and $N(3, 4)=4$. Using Equation~\ref{PMI}, we got PMI($3, 4$)=0.058. 
As described in~\cite{yao2019graph}, a positive PMI value indicates a high semantic correlation of words in
corpus and vice versa.

%The PMI score only catches the word pair co-occurrence information. For the document-word information, the term frequency-inverse document frequency (TF-IDF) of the word in the document is considered. 
The adjacency matrix $\textbf{\textit{A}}$ of the graph is then computed elementwise, as follows:
%Thus the adjacency matrix, which stores the weights of the edges, is computed as follow:
\begin{equation} \label{adj}
\small
A_{i, j}=\left\{\begin{matrix}
\textrm{PMI}(i, j) &i, j \textrm{ are word nodes, } \textrm{PMI}(i, j)>0; \\ 
& n_D< i,j \leqslant n_D + n_W \\
\textrm{TF-IDF}_{i,j} & \textrm{document node } i \textrm{ and word node } j;   \\ 
& i\leqslant n_D; n_D < j \leqslant n_D +n_W\\
1 & i=j \\ 
0 & \textrm{otherwise}
\end{matrix}\right.
\end{equation}   

Since the graph is undirected, the adjacency matrix is symmetric. Finally, the adjacency matrix is normalized by $\tildeb{\textbf{\textit{A}}} = \textbf{\textit{D}}^{-\frac{1}{2}}\textbf{\textit{A}}\textbf{\textit{D}}^{-\frac{1}{2}}$, where $\textbf{\textit{D}}$ is the degree matrix of $\textbf{\textit{A}}$. The normalized adjacency matrix $\tildeb{\textbf{\textit{A}}}$ is used to weight the graph node features, cf.~Section~\ref{smm}. A PyTorch implementation based on Text-GCN~\cite{yao2019graph}, as provided on GitHub\footnote{\url{https://github.com/codeKgu/Text-GCN}}, was used for the implementation.

\section{System Overview}\label{Overview}
In this section, we specify our
uni- and bi-modal models. 
\begin{figure}[!htb]
     \centering
     \hspace{-0.3cm}
     \begin{subfigure}[b]{0.123\textwidth}
         \centering
         \includegraphics[scale=0.515]{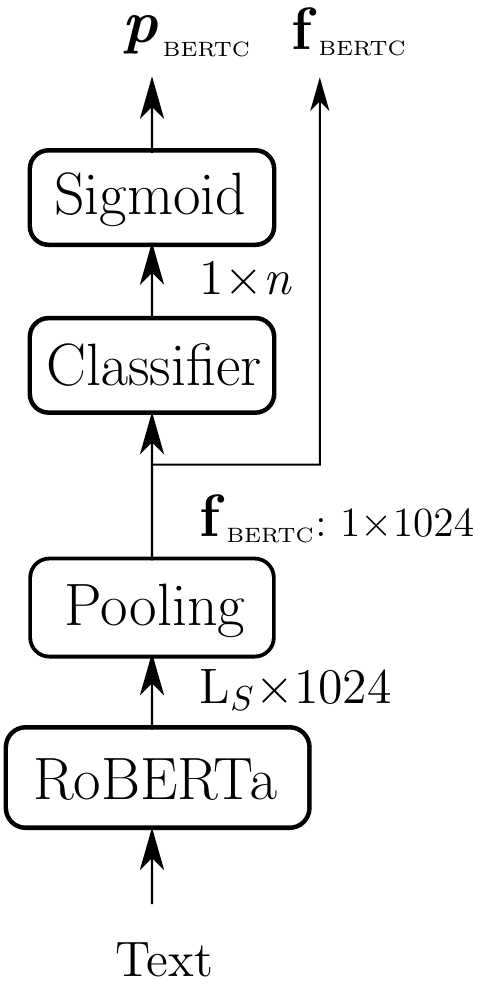}
         \caption{BERTC}
         \label{BERTC}
     \end{subfigure}
     \hspace{0.5cm}
     \begin{subfigure}[b]{0.123\textwidth}
         \centering
         \includegraphics[scale=0.515]{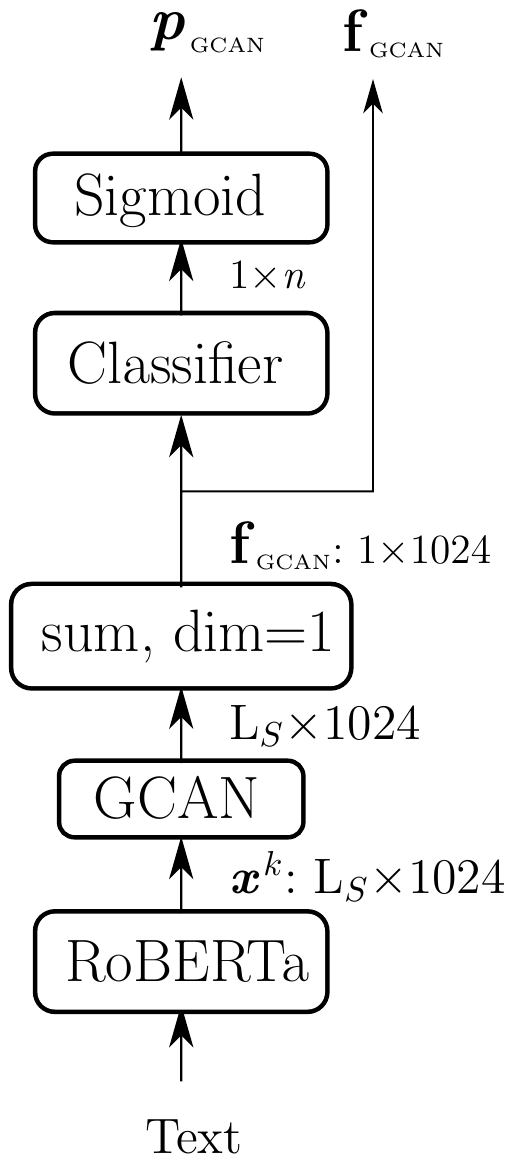}
         \caption{GCAN}
         \label{GCAN}
     \end{subfigure}
     \hspace{0.65cm}
     \begin{subfigure}[b]{0.123\textwidth}
         \centering
         \includegraphics[scale=0.515]{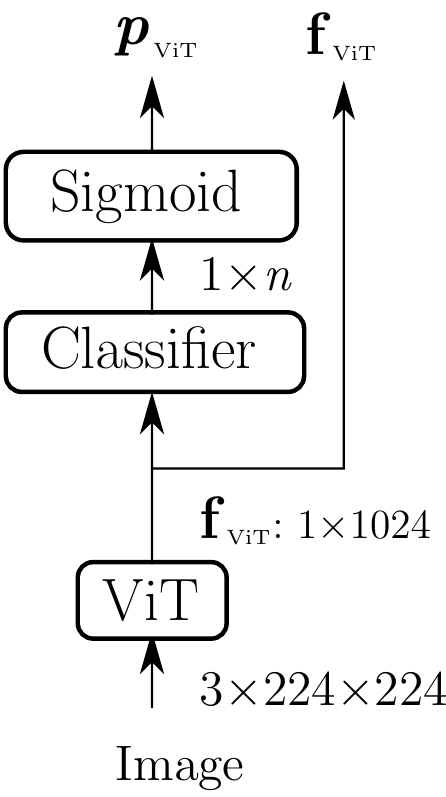}
         \caption{ViT}
         \label{Image}
     \end{subfigure}
        \caption{Uni-modal models, where $\textrm{L}_S$ is the sequence length, which depends on the RoBERTa tokenizer.}
        \label{fig:smms}
\end{figure}

% Proposal: In the following we outline our architectures. We first present our single- and multi modal models. Then we first ensemble all trained versions of a architecture that we have and subsequently the make a final ensemble out of the ensembles of different architectures.

Figure~\ref{fig:smms} depicts the three uni-modal models BERTC (\ref{BERTC}), GCAN (\ref{GCAN}), and ViT (\ref{Image}), which form the basis of our further experiments. The bi-modal models are constructed based on trained uni-modal models and our proposed model fusion network, which is further detailed in Section \ref{mmm}. Finally, we apply soft and hard voting ensembles on the trained candidate models.

\subsection{Uni-modal models}\label{smm}
As illustrated  in Figure~\ref{fig:smms}, every uni-modal model has two outputs: the classification probabilities $\textbf{\textit{p}}_{\textrm{\tiny{i}}}$ and the classification features $\textrm{\textbf{f}}_{\textrm{\tiny{i}}}$. All classifier blocks in our models have the following, identical structure: a fully connected layer reduces the feature dimension to half the input dimension, followed by a ReLU activation and a dropout layer. Ultimately, an output layer projects the features to the output dimension $n$, and a sigmoid function squashes the range of the output vector components to $(0,1)$, allowing for an interpretation as a vector of label probabilities, with possible overlap in categories.

\noindent \textbf{BERTC}: We fine-tune a pre-trained large RoBERTa language model (\texttt{roberta-large}) for classification. The text input is encoded by the RoBERTa model with the embedding dimension 1024. The Pooler layer returns the first classification token \texttt{[cls]} embedding $\textrm{\textbf{f}}_{\textrm{\tiny{BERTC}}}$ and feeds it into the classifier to obtain the probabilities $\textbf{\textit{p}}_{\textrm{\tiny{BERTC}}}$.

\noindent \textbf{GCAN}: Again, a pre-trained RoBERTa model extracts contextual text information. Each token is considered as a word node and each meme is a document node. Thus the word node representation is given by the corresponding RoBERTa word embedding vector. 
% Proposal: For this model we again use the pre-trained RoBERTa model to extract a contextual text representation. 
% We call this embedding of document k x_k = [...] where each each component x_k_i is a 1024 embedding of the i-th token. The embedding thus is a Slen x 1024 matrix. The first token x_k_1 is the [cls] token and is an embedding of the entire document, the rest of the tokens are contextal word embeddings. Thus we can construct the graph G where each token is a node and its embedding are the attached features. The edge weights are computed as presented above. Additionally our GCAN block adopts a multihead self attention mechanism in each GCAN layer to transform the node features...
We denote the input embedding sequence of document $k$ as $\textbf{\textit{x}}^k=[ \textbf{\textit{x}}_1^k, \textbf{\textit{x}}_2^k, \cdots \textbf{\textit{x}}_{\textrm{L}_S}^k]$, where $\textbf{\textit{x}}_i^k$, $i\in \{1, \ldots, \textrm{L}_S\}$ is a 1024-dimensional embedding vector of the $i$-th token. As depicted in Figure~\ref{GCAN}, $\textbf{\textit{x}}^k$ is an $\textrm{L}_S\times$1024 matrix. The first classification token \texttt{[cls]} embedding represents the document classification information. Thus, we use the document-word co-occurrence information TF-IDF as the edge weights for the \texttt{[cls]} embedding. All other token embeddings are weighted with the word co-occurrence information PMI. 

For each document $k$, we extract its specific adjacency matrix ${\tildeb{\textbf{\textit{A}}}}_k$ from the complete adjacency matrix ${\tildeb{\textbf{\textit{A}}}}$ by reducing it to rows and columns of all the document and word nodes ($i$ and $j$ in Equation~\ref{adj}) that are present in this document.  
%we take the sub adjacency matrix ${\tilde{\textbf{\textit{A}}}}'_k$ %from the whole  adjacency matrix $\tilde{\textbf{\textit{A}}}$ according to the document and word nodes of the current document. 
The extracted document adjacency matrix ${\tildeb{\textbf{\textit{A}}}}_k$ is an $\textrm{L}_S\times\textrm{L}_S$ matrix.

%\todo[inline]{ich verstehe den Zusammenhang mit den Gleichungen (1)-(4) nicht. Mindestens bitte die Gleichungen hier noch mal referenzieren. Die Konstruktion der Document adjacency matrix könntest Du evtl. besser oben im Abschnitt zum GCAN erklären}
    
The GCAN block in Figure~\ref{GCAN} adopts the multi-head self-attention mechanism in 3 successive GCAN layers 
to embed the node representations. The queries \textbf{Q}, keys \textbf{K} and values \textbf{V} are identical and set to the respective layer input. 
The first layer input is given by the RoBERTa word embeddings $\textbf{\textit{x}}^k$ of the input text. The attention of head $j$ is obtained by
\vspace{-0.2cm}
\begin{equation} \label{dotatttransform}
    \small
    \bm{\alpha}_j = \textrm{softmax}\left(\frac{\left(\textbf{W}_j^Q\textbf{Q}^T\right)^T\left(\textbf{W}_j^K\textbf{K}^T\right)}{\sqrt{d_k}}\right)\left(\textbf{W}_j^V\textbf{V}^T\right)^T
\end{equation}

where $\textbf{W}_j^\ast$ are learned parameters for the queries $\textbf{Q}$, keys $\textbf{K}$, and values $\textbf{V}$, respectively. A superscript $T$ denotes the transpose; $d_k=\frac{d_{att}}{h}$, $d_{att}$ is the attention dimension and $h$ is the number of attention heads. Having computed the multi-head self-attention, each attention head output is multiplied by the document adjacency matrix ${\tildeb{\textbf{\textit{A}}}}_k$
\begin{equation} \label{dotatt}
    \tildea{\bm{{\alpha}}}_j = {\tildeb{\textbf{\textit{A}}}}_k\bm{\alpha}_j.
\end{equation}

Equation~\ref{layerprojection} describes the output $\bm{\alpha}$ of the GCAN layer: The weighted outcomes all heads are concatenated (concat), and a fully connected layer (FC) projects the representation to the attention dimension. Inspired by~\cite{velivckovic2017graph}, instead of concatenating the weighted attention head outputs, we employ averaging (avg) to fuse these weighted outputs in the last GCAN layer. A fully connected layer again projects the final representation to the attention dimension. Thus, after the GCAN block, the text  representation is still an $\textrm{L}_S\times$1024 matrix. The document classification feature vector $\textrm{\textbf{f}}_{\textrm{\tiny{GCAN}}}$ is obtained by summing all node  representations.
\begin{equation} \label{layerprojection}
\small
\bm{\alpha}=\left\{\begin{matrix}
 \textrm{FC}\left ( \textrm{concat}\left ( \tildea{\bm{{\alpha}}}_1, \cdots \tildea{\bm{{\alpha}}}_h  \right ) \right ) & \textrm{not in last layer} \\
\textrm{FC}\left ( \textrm{avg}\left ( \tildea{\bm{{\alpha}}}_1, \cdots \tildea{\bm{{\alpha}}}_h  \right ) \right ) & \textrm{in last layer}  \\
\end{matrix}\right.
\end{equation}  

\noindent \textbf{ViT}: To extract the visual contextual information, we utilize the pre-trained ViT model  \texttt{vit-large-patch16-224} to encode the input image. For this purpose, the input image is split into fixed-size patches, and a linear projection of the flattened patches is used to obtain the patch embedding vectors. The Transformer encoder transforms the embedding vectors. Finally, the  embedding  $\textrm{\textbf{f}}_{\textrm{\tiny{ViT}}}$ of the first classification token, \texttt{[cls]}, is fed to the classifier to obtain the prediction probabilities $\textbf{\textit{p}}_{\textrm{\tiny{ViT}}}$.

\subsection{Bi-modal models}\label{mmm}
% Proposal: Each of our single-modal models has two outputs, the classification probabilities $\textbf{\textit{p}}_{i}$ and the classification features classification features $\textrm{\textbf{f}}_{i}$. We now combine the outputs of those models to an ensemble as depicted in Figure~\ref{fig:mmms}. 
% To combine the inputs we use two fusion strategies: the stream weighting decision fusion and the representation fusion. The stream weights are computed by a fully connected network from the classification features. The output of each model is then weighted and summed to produce $\textbf{\textit{p}}_{sw}=\sum_{i}^{}\textbf{\textit{p}}_i\cdot w_i$. The representation fusion computes a new classification probability from the classification features of every participating model. Finally, the two new predictions are averaged to produce a final probability. For this challenge we considered the following model combinations, where each single model was pre-trained and now finetuned. [now the table]

\begin{figure}[!htb]
        \centering
        %\hspace{-0.5cm}
        \includegraphics[scale=0.50]{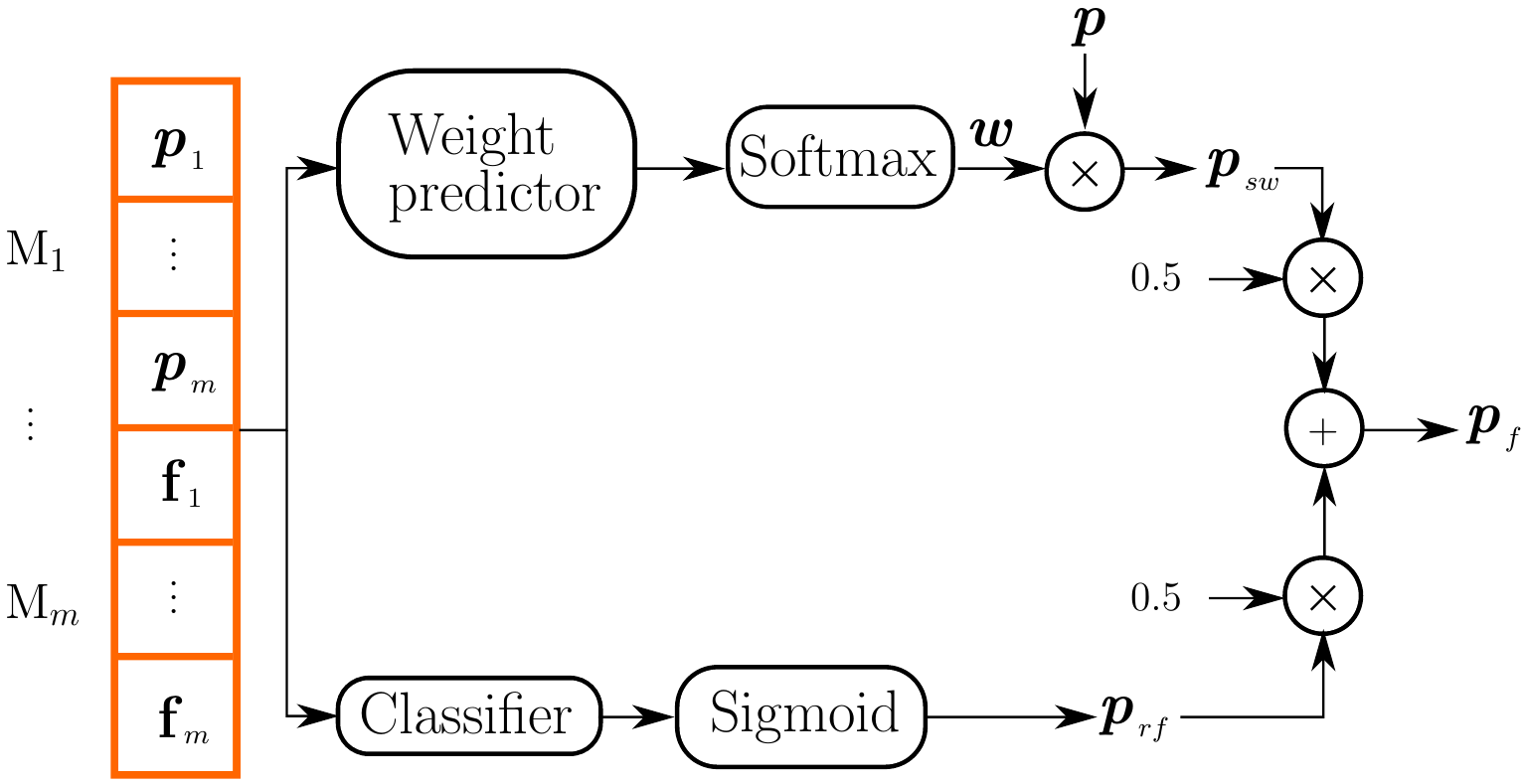}
         \caption{Fusion model structure}
         \label{fig:mmms}
\end{figure}

Figure~\ref{fig:mmms} shows our fusion model structure. 

Each model $\textrm{M}_i$ has two outputs: the vector of its classification probabilities $\textbf{\textit{p}}_{i}$ and the classification features $\textrm{\textbf{f}}_{i}$. We concatenate the model classification probabilities and features as a multi-modal representation to make the final decision. 

Two fusion strategies---stream-weighting-based decision fusion and representation fusion---are considered. The weight predictor and the classifier in Figure~\ref{fig:mmms} both have the same structure as the classifier block in Figure~\ref{fig:smms}. The weight predictor output dimension is the number  $m$ of candidate models for fusion. The stream weighting probability $\textbf{\textit{p}}_{sw}$ is obtained through a weighted combination of the class probability vectors of all uni-modal model outcome probabilities, i.e.~
\begin{equation}\textbf{\textit{p}}_{sw}=\sum_{i}^{}\textbf{\textit{p}}_i\cdot w_i.
\end{equation}

The classifier output dimension is the same as the number of classes $n$. A sigmoid function computes the representation fusion probabilities $\textbf{\textit{p}}_{rf}$ from the combined multi-modal representation. Finally, we average the stream weighting and the representation fusion probabilities. The following model combinations are attempted, where $\textrm{M}_i$, $i\in{\{1, 2, 3\}}$ is the $i$-th pre-trained uni-modal model.

\renewcommand{\arraystretch}{1.15}
\begin{table}[H]
\setlength{\tabcolsep}{6pt} % Default value: 6pt
\begin{tabular}{|c|c c c|}
\hline
\textbf{Bi-modal model}               & $\textrm{M}_1$ & $\textrm{M}_2$ & $\textrm{M}_3$ \\ \hline\hline
BERTC-ViT      & BERTC  & ViT    & -      \\ 
GCAN-ViT       & GCAN   & ViT    & -      \\ 
BERTC-GCAN     & BERTC  & GCAN   & -      \\ 
BERTC-GCAN-ViT & BERTC  & GCAN   & ViT    \\ \hline
\end{tabular}
\end{table}

\subsection{Ensemble learning}\label{ensemble}
Having established a number of possible uni-modal and bi-modal models, we now combine these trained models into ensembles. It has been reported in many studies that ensemble learning can enhance performance in comparison to single learners~\cite{onan2016ensemble, zhu2020enhance, gomes2017survey}. Therefore, we consider soft and hard voting ensemble approaches.

%Proposal: Since we train each model with 10-fold cross-validation, we have 10 versions of these models. In the first ensemble stage, the dataset-level, we combine those in a soft voting approach.

We use the Python \texttt{sklearn} package\footnote{\url{https://github.com/scikit-learn/scikit-learn}} for 10-fold cross-validation.  %In 10-fold cross-validation, the training set is first divided into ten subsets. In each fold, the model selects one subset as the inner-test set, and the rest nine subsets are the training set. 
Thus, each model structure was trained ten times with different inner test sets. Finally, these ten models are used to evaluate the official test set and deliver ten predictions for every sample. The soft voting ensemble method is implemented as follows:  $\textbf{\textit{p}}_{\small \textrm{M}_i}$, the ensemble probabilities that are used in the overall class decisions, are computed via
\begin{equation} \label{softvote}
    \textbf{\textit{p}}_{\small \textrm{M}_i}=\sum_{j=0}^{9}w_{\small \textrm{M}_i}^j\cdot \textbf{\textit{p}}_{\small \textrm{M}_i}^j.
\end{equation}
Here, $\textbf{\textit{p}}_{\small \textrm{M}_i}^j$ denotes the probabilities of model $\textrm{M}_i$ in the $j$-th fold. The weights $w_{\small \textrm{M}_i}^j$ are computed by 
\begin{equation} \label{weight}
w_{\small \textrm{M}_i}^j = \frac{\textrm{F1}_{\small \textrm{M}_i}^j}{\sum_{f}^{}\textrm{F1}_{\small \textrm{M}_i}^f}.
\end{equation}
$\textrm{F1}_{\small \textrm{M}_i}^j$ corresponds with the  best F1-score of model $\textrm{M}_i$ over all epochs, computed on the inner test set in fold $j$.  This soft voting ensemble, using the same model structure, but with the multiple outcomes from 10-fold cross-validation, is referred to as a \emph{dataset-level ensemble} in the following.

% Proposal: Having an ensemble of each architecture among the cross validations, we ensemble different architectures with hard fusion. The result of that are our final candidates.

The second type of ensemble---the \emph{model-level ensemble}---is constructed from the \emph{dataset-level ensemble} results of each model. We use a hard voting strategy with seven candidate models (BERTC, GCAN, ViT, BERTC-ViT, GCAN-ViT, BERTC-GCAN, and BERTC-GCAN-ViT). In this approach, we set the final prediction for a data point to one, if at least half of the considered models vote one, making it a simple majority-voting strategy.

\section{Experimental Setup}\label{setup}
% Proposal: The challenge dataset provides a transcription text stream that was obtained via OCR. Via image captioning we derive a second text stream that contains a description of the image in a few words.
% Since the transcription is not manually derived it contains unreadable sections where non latin characters are parsed. We remove those with the Python \texttt{ftfy} package\footnote{\url{https://github.com/rspeer/python-ftfy}}.

In this section, we describe our data processing and training pipeline in more detail.
\subsection{Data pre-processing}\label{datapreprocessing}
%Two kinds of text data are used in our work. One is the OCR text transcriptions of memes, as provided by the organizers. The second is our self-produced image caption text, which describes the image with a few words.
The challenge dataset provides a transcription text stream that was obtained via OCR. Via image captioning, we derive a second text stream that contains a description of the image in a few words. 

For the OCR text, we first use the Python \texttt{ftfy} package\footnote{\url{https://github.com/rspeer/python-ftfy}} to fix the garbled sequences that result from unexpected encodings (the \emph{mojibake}) like "à¶´à¶§à·". Next, all "@", "\#" symbols and website addresses are removed from the text. The emojis are converted to text form by the Python \texttt{emoji} package\footnote{\url{https://github.com/carpedm20/emoji}}. Finally, we remove non-English characters and convert the text to lowercase. 

\begin{figure}[H]
     \centering
     \begin{subfigure}[b]{0.15\textwidth}
         \centering
         \includegraphics[scale=0.15]{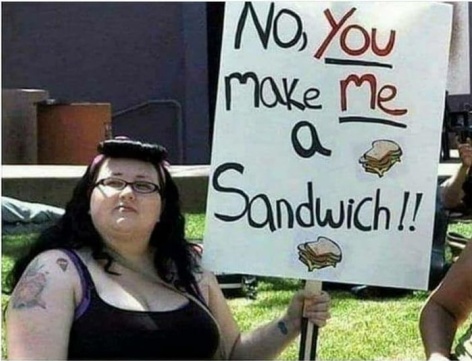}
         \caption{a woman holding a sign in front of a sign}
         \label{first}
     \end{subfigure}
     \hfill
     \begin{subfigure}[b]{0.15\textwidth}
         \centering
         \includegraphics[scale=0.15]{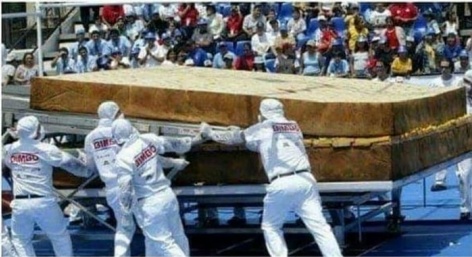}
         \caption{a couple of baseball players standing next to each other}
         \label{second}
     \end{subfigure}
     \hfill
     \begin{subfigure}[b]{0.15\textwidth}
         \centering
         \includegraphics[scale=0.1]{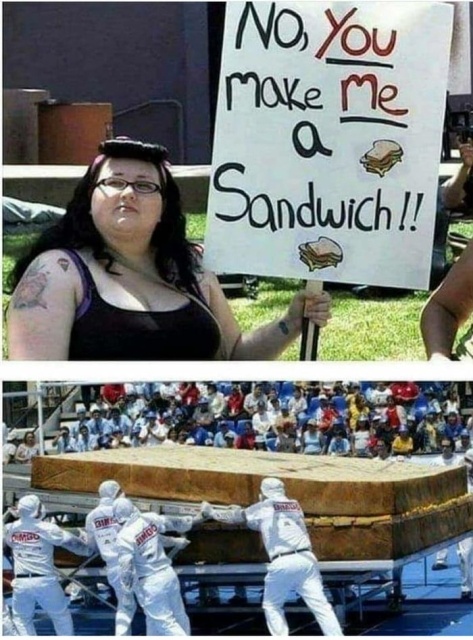}
         \caption{a woman standing next to a group of people}
         \label{org}
     \end{subfigure}
        \caption{In (a) and (b), we see "sub-images" and corresponding captions. (c) shows the meme and its caption (when not considering the sub-image structure).}
        \label{fig:meme}
\end{figure}

For image captioning, we  utilize a pre-trained encoder-decoder attention model~\cite{xu2016show}\footnote{\url{https://github.com/sgrvinod/a-PyTorch-Tutorial-to-Image-Captioning}}. %, which is based on the encoder-decoder structure and trained on the MSCOCO\textquotesingle 14 dataset. 
Although the translation from image to text is not very accurate, most likely owing to issues like the overlaid meme text, it was nonetheless beneficial for our classification task. We found that the description becomes more precise, when we split the memes into their constituent sub-images where applicable. In that case, the image caption is extracted over every sub-image as well as the entire meme. Finally, the image captions are combined with the word "and" and then concatenated with the OCR text, separated by ". ". With this rule, the final text of the meme in Figure~\ref{fig:meme} is: "\textit{mgo ci  aindo  make make me  sandwich!!. a couple of baseball players standing next to each other and a woman holding a sign in front of a sign and a woman standing next to a group of people.}" 

We use the entire meme as the image input for ViT. All memes are first resized to 256$\times$256 and center-cropped to 224$\times$224 dimensions. The ViT model uses all 3 RGB channels, so we retain the RGB structure, thus the input image dimension is  3$\times$224$\times$224. We regularize the entire image database to range 0 to 1, then normalize each individual image to have zero mean and unit variance.

\subsection{Loss function }\label{weightloss}
We decided to use the binary cross-entropy (BCE) loss for both subtasks. 

Due to the imbalance in the class distributions (see Table~\ref{tab:datadis}), in sub-task B, we weighted the class-specific loss terms by their support as follows:
\begin{equation} \label{lossweight}
    w_{c} = \frac{\frac{\textrm{NoS}}{\textrm{NoS}(c)}}{\sum_{c'}^{}\frac{\textrm{NoS}}{\textrm{NoS}(c')}}, c \in [\textrm{Shm}, \textrm{Ste}, \textrm{Obj}, \textrm{Vio}]
\end{equation}
where $\textrm{NoS}$ is the total number of samples in the training set and $\textrm{NoS}(c)$ represents the number of true instances for class $c$. 
%Only four categories are considered in sub-task B: shaming, stereotype, objectification, and violence.
The loss is then computed through the weighted combination of the single BCE terms:
\begin{equation} \label{loss}
\mathcal{L}_1 = \sum_{c}^{}w_{c}\cdot \textrm{BCE}(\textbf{\textit{p}}_c^B, \textbf{\textit{y}}_c^B).
\end{equation}
Here, $\textbf{\textit{p}}_c^B$ represents the system's output probability of class $c$ and $\textbf{\textit{y}}_c^B$ is the binary ground truth for sub-task B. 

Additionally, we employ a teacher forcing loss to connect both subtasks. The idea is that an instance should be identified as misogynous and possibly grouped into sub-categories simultaneously. The teacher forcing loss is defined as:

\begin{equation} \label{mse}
\mathcal{L}_2 =
\lVert \textbf{\textit{p}}^A - \textbf{\textit{y}}^A 
\rVert, 
\end{equation}
where the system's output probability for sub-task A is determined as:
\begin{equation} \label{mse1}
\textbf{\textit{p}}^A = \textrm{max}\big(\textbf{\textit{p}}_{_\textrm{Shm}}^B, \textbf{\textit{p}}_{_\textrm{Ste}}^B, \textbf{\textit{p}}_{_\textrm{Obj}}^B, \textbf{\textit{p}}_{_\textrm{Vio}}^B \big).
\end{equation}
The final loss is computed by
\begin{equation} \label{lossend}
\mathcal{L} = 0.7\cdot \mathcal{L}_1 + 0.3 \cdot \mathcal{L}_2.
\end{equation}

\subsection{Model training}\label{others}
 All models are trained using the PyTorch library~\cite{paszke2019pytorch} for 50 epochs. The AdamW optimizer~\cite{loshchilov2017decoupled} is used for backpropagation, using a linear learning rate scheduler with a warm-up to adapt the learning rate during the first four epochs in the training stage. The dropout rate is 0.5. The RoBERTa model parameters in the BERTC and the GCAN model are optimized separately. 

In our GCAN model, the adjacency matrix is computed with a sliding window of length 10. An 8-head self-attention is applied over 3 GCAN layers with an attention dimension of 1024.

For all uni-modal models, the batch size is 16 and the initial learning rate is $2\cdot10^{-5}$. The RoBERTa and ViT block parameters in Figure~\ref{fig:smms} are also fine-tuned. The bi-modal models are trained based on the pre-trained uni-modal models. Here, we choose the batch size as 32, the initial learning rate is \mbox{$5\cdot10^{-6}$}. As the RoBERTa and ViT block parameters in Figure~\ref{fig:smms} are already updated during the uni-modal training stage, we froze these parameters in bi-modal re-training. 

To avoid overfitting, we adopt early stopping to exit the training process when the computed F1-score on the inner test set does not increase over 4 epochs. Inspired by~\cite{DBLP:conf/iclr/HuangLP0HW17}, we finally averaged those two epoch-wise model parameters,  which had the highest validation F1-score during the training stage. 

The models have the same structure for sub-tasks A and B. The only differences are that in sub-task A, the classifier output dimension $n$ is $1$, and the BCE is used as the loss function (\emph{Setup A}), whereas in sub-task B, the classifier output dimension $n$ equals $4$ and training uses the weighted BCE with teacher forcing (Equation~\ref{lossend}) as the loss function (\emph{Setup B}). All models are trained using NVIDIA’s Volta-based DGX-1 multi-GPU system, using 3 Tesla V100 GPUs with 32~GB memory each.

\section{Results}\label{results}
In summary, we investigated two configurations, displayed in Table~\ref{summary}. \emph{Setup A} represents the binary classification for sub-task A, resulting in an output dimension $n=1$. \emph{Setup B} additionally deals with the  multi-label classification of sub-task B, returning an output of dimension $n=4$. All results are evaluated on the official test set.

\begin{table}[!htb]
\setlength{\tabcolsep}{2.7pt} % Default value: 6pt
\begin{tabular}{|c|c|c|c|}
\hline
\textbf{Setup}   & \textbf{Task}       & \textbf{Dimension} & \textbf{Loss}         \\ \hline\hline
\emph{Setup A} & sub-task A  & $n=1$          & BCE          \\ \hline
\emph{Setup B} & \small{sub-tasks A/B} & $n=4$          & {\begin{tabular}[c]{@{}c@{}}\small{Weighted BCE} \& \\\small{Teacher Forcing} \end{tabular}} \\ \hline
\end{tabular}
\caption{Summary of the considered configurations.}
\label{summary}
\end{table}

%There are two model setups (\emph{setup-A} and \emph{setup-B}) for
%We build two ensemble learning models for both 
%sub-task A and B. All models are evaluated on the official test set. %Our experimental results show that computing sub-task A results with \emph{setup-B} has better performance. %Therefore, we only analyze here the results from \emph{setup-B} for both sub-tasks. The detailed \emph{setup-A} results are described in Appendix~\ref{sub-taska}

\subsection{Results for \emph{Setup A} (Sub-task A)}
\label{sub-taska}
\begin{figure}[!htb]
        \centering
        \includegraphics[scale=0.5]{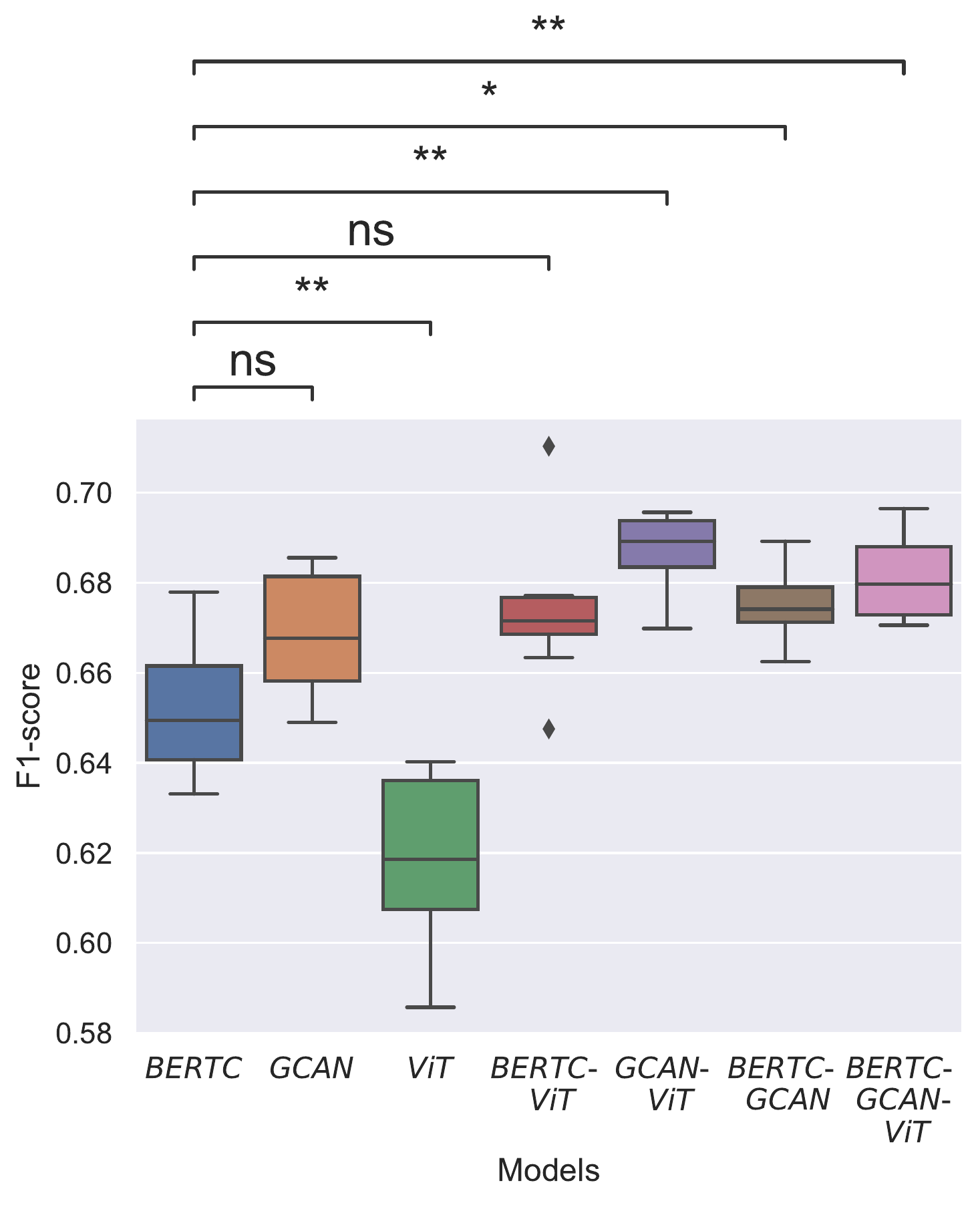}
         \caption{Macro-average F1-scores for sub-task A based on 10-fold cross validation. Asterisks indicate a statistically significant
difference, where ** denotes 1e-04 < p <= 1e-03, * corresponds to 1e-02 < p <= 5e-02, and ns indicates results where p > 5e-02.}
         \label{fig:subtaskacv}
\end{figure}

In the first stage, we trained three different uni-modal models (i.e., BERTC, GCAN, and ViT). In the second stage, we optimized the  bi-modal models (i.e., BERTC-ViT, GCAN-ViT, BERTC-GCAN, and BERTC-GCAN-ViT).  
The evaluation results in terms of the macro-average F1-score are displayed in Figure~\ref{fig:subtaskacv} and Table~\ref{subtaskacv}, showing the performance in identifying misogynous memes. 
To assess the statistical significance of performance differences, we applied a 10-fold cross validation and computed the Mann-Whitney-U test~\cite{mann1947test}. 

As we can see, the text-only models (BERTC and GCAN) generally show a superior performance compared to the image-only model (ViT). 
%We have also tried other image models like the EfficientNet~\cite{tan2019efficientnet}, but the ViT model outperforms. %In all single-modal models, the GCAN model is superior to other models.  
The results in Figure~\ref{fig:subtaskacv} clearly indicate robust performance for our bi-modal models. They are more accurate and robust. In summary, the GCAN-ViT model yields the best results w.r.t. the reported median F1-score.

\begin{table}[!htb]
\setlength{\tabcolsep}{1.5pt} % Default value: 6pt

\begin{tabular}{|c|c||c|c|}

\hline
\textbf{Model}     & \textbf{Ensemble} & \textbf{Model}          & \textbf{Ensemble} \\ \hline\hline
BERTC     & 0.663            & GCAN-ViT                                                  & \textbf{0.707}            \\ \hline
GCAN      & 0.674            & \small{BERTC-GCAN}                                                & 0.677            \\ \hline
ViT       & 0.619            & \begin{tabular}[c]{@{}c@{}}\small{BERTC-}\\ \small{GCAN-ViT}\end{tabular} & 0.689            \\ \hline
BERTC-ViT & 0.697            & -                                                         & -                \\ \hline
\end{tabular}
\caption{Macro-average F1-scores of soft voting ensembles for sub-task A.}
\label{subtaskacv}
\end{table}

Table~\ref{subtaskacv} lists the averaged F1-scores for soft voting ensembles, obtained by combining all learned models from the 10-fold cross-validations. The results show that our GCAN-ViT model outperforms all other models, achieving an F1-score of 0.707.

%Finally, the model-level hard voting ensemble is adopted and yields 0.698 F1-score. However, our models for sub-task A suffered from mismatch on the official test set. In the early research stage, we worked only on the training set with 10,000 samples. 700 samples were randomly selected and split from the training set as a temporary test set. The rest 9,300 samples were used as the training set to train our proposed models, 10-fold cross-validation was still applied. Our ensemble results on the temporary test set were over 0.87. It could be due to a higher level of similarity between the training set and the temporary test set.

\subsection{Results for \emph{Setup B} (Sub-tasks A/B)}\label{sub-taskb}
\begin{figure*}[!htb]
     \centering
     \begin{subfigure}[b]{0.4\textwidth} %{0.1\textwidth}
         \centering
         \includegraphics[scale=0.5]{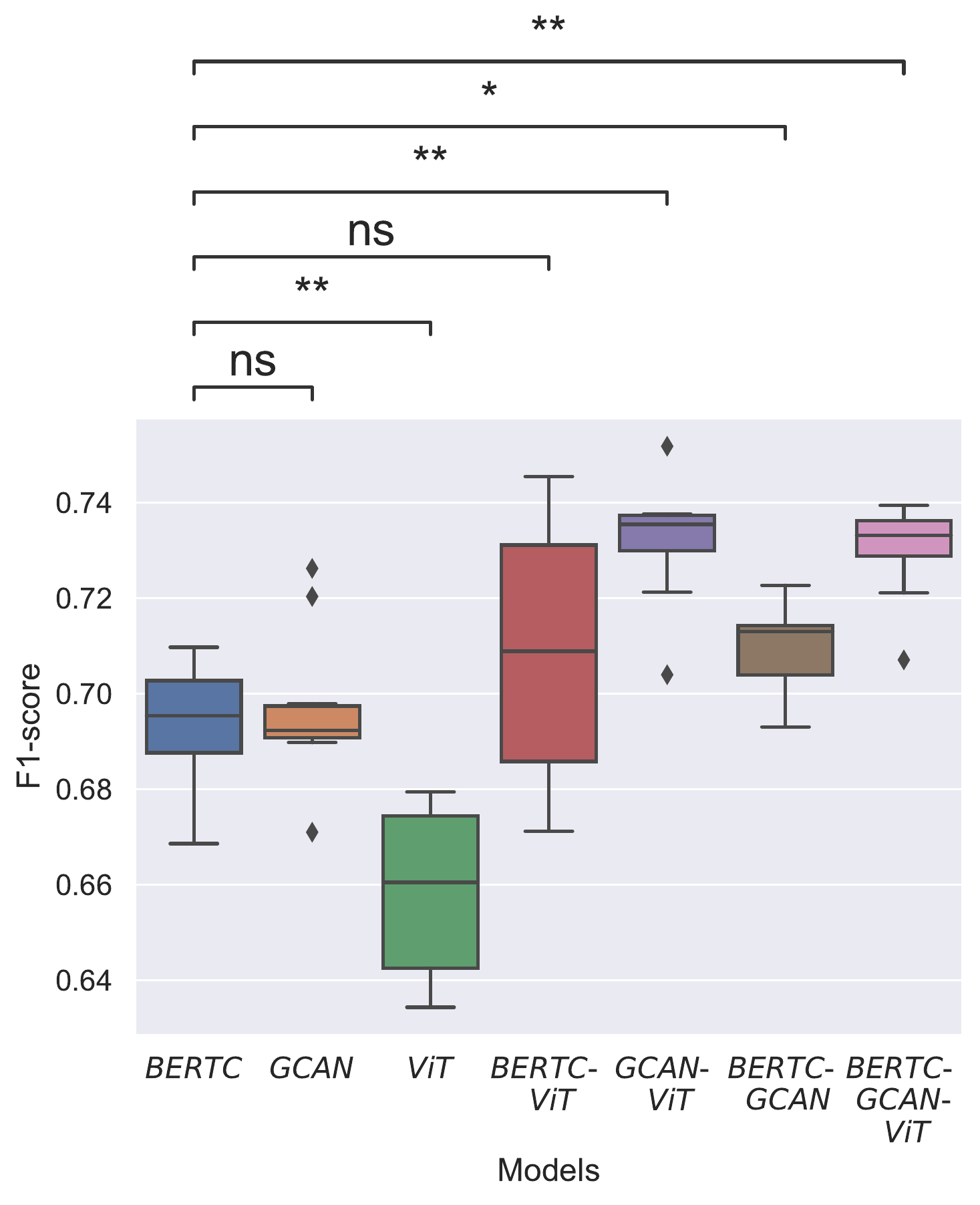}
         \caption{Results for sub-task A.}
         \label{fig:sigA}
     \end{subfigure}
     \hspace{1.8cm}%\hfill
     \begin{subfigure}[b]{0.4\textwidth} %{0.1\textwidth}
         \centering
         \includegraphics[scale=0.5]{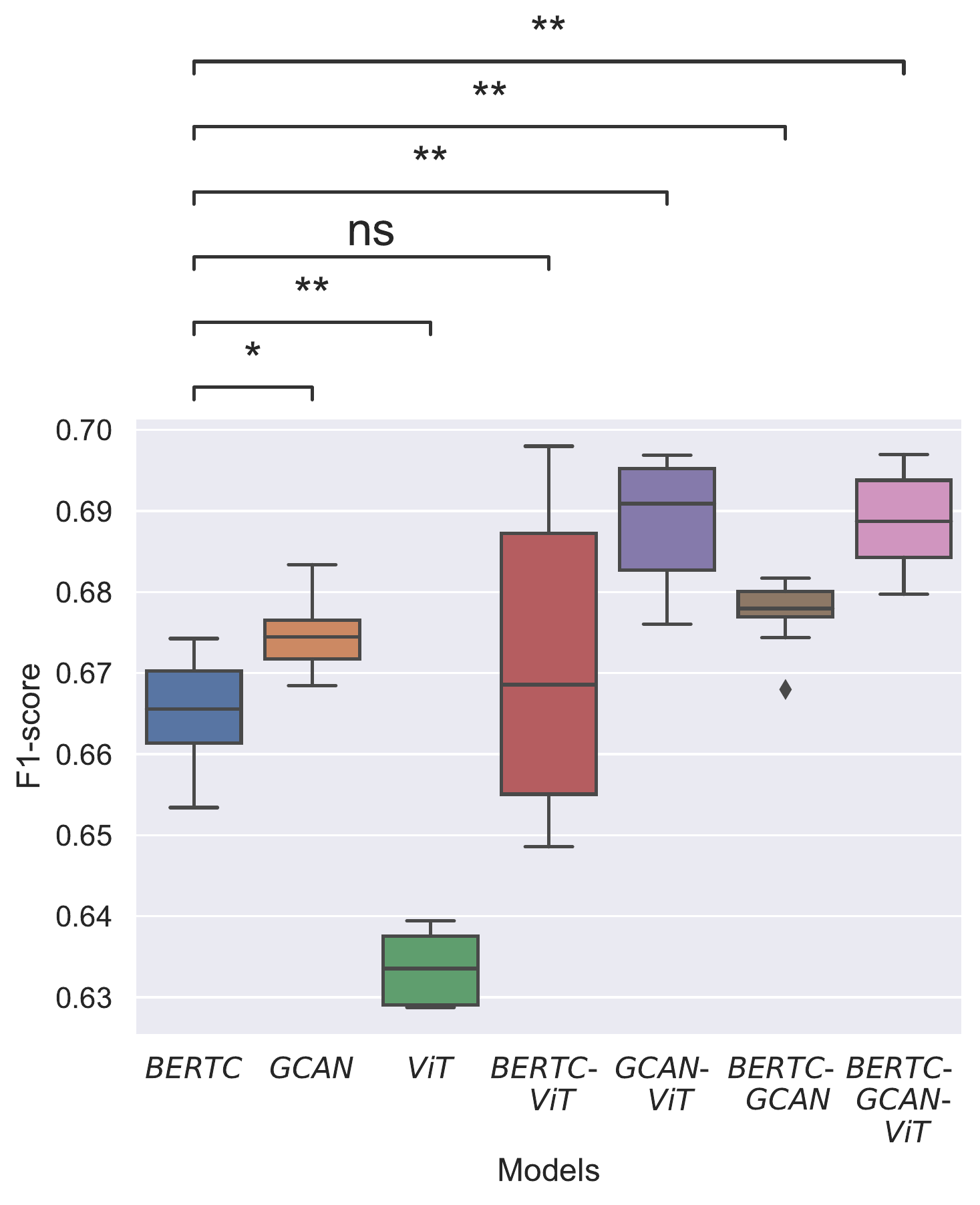}
         \caption{Results for sub-task B.}
         \label{fig:sigB}
     \end{subfigure}
        \caption{Performance for \emph{Setup B}. The notation is defined in Figure~\ref{fig:subtaskacv}.}
        \label{fig:sigs}
\end{figure*}

Next, we addressed sub-task B, i.e.~to classify the misogynous memes into four, potentially overlapping, categories. 
Similar to \emph{Setup A}, we trained the same uni- and bi-modal models, but incorporating a different loss (see Table~\ref{summary}).
For sub-task B, the weighted-average F1-score is applied. The results are presented in Figure~\ref{fig:sigs}.

%\begin{equation} \label{f1weight}
%\begin{matrix}
%\textrm{weighted F1-score} = \sum_{i}^{}\frac{\textrm{NoS}(i)}{ %\textrm{NoS}}\cdot \textrm{F1-score}_i, \\
%i \in [\textrm{Shm}, \textrm{Ste}, \textrm{Obj}, \textrm{Vio}],
%\end{matrix}
%\end{equation}
%is used in the evaluation of Sub-task B.

Interestingly, the models optimized for sub-task B also perform better for sub-task A. In this case, we set the estimated label "misogynous" to 1 if at least  one of the labels for "shaming", "stereotype", "objectification", or "violence" is 1. 

Figure~\ref{fig:sigA} depicts the sub-task A results while Figure~\ref{fig:sigB} shows the corresponding performance for sub-task B.  Again, we see that the bi-modal model GCAN-ViT outperforms all other models.

In addition, Tables~\ref{subtaskabcv} and~\ref{subtaskall} show the results for soft and hard voting ensembles. By comparing Table~\ref{subtaskabcv} with Table~\ref{subtaskacv} (both tables represent soft voting results), we observe  significantly improved F1-scores for \emph{Setup B}. 

\begin{table}[!htb]
\setlength{\tabcolsep}{5pt} % Default value: 6pt

\begin{tabular}{|c|c|c|}
\hline
\textbf{Model}     & \textbf{Sub-task A} & \textbf{Sub-task B}     \\ \hline\hline
BERTC     & 0.714        & 0.684            \\ \hline
GCAN      & 0.725           & 0.695        \\ \hline
ViT       & 0.666             & 0.641        \\ \hline
BERTC-ViT & 0.746          &0.692               \\ \hline
GCAN-ViT  & \textbf{0.758}    & \textbf{0.704}       \\ \hline
BERTC-GCAN & 0.724           & 0.696   \\ \hline
BERTC-GCAN-ViT & 0.755         & 0.704       \\ \hline

\end{tabular}
\caption{F1-scores of soft voting ensembles for \emph{Setup B} (sub-tasks A and B).}
\label{subtaskabcv}
\end{table}

%\begin{table}
%\setlength{\tabcolsep}{3pt} % Default value: 6pt

%\begin{tabular}{|c|c|c|c|}
%\hline
%model     & ensemble & model          & ensemble \\ \hline
%BERTC     & 0.684           & GCAN-Vit                                                  & \textbf{0.704}            \\ \hline
%GCAN      & 0.695            & BERTC-GCAN                                                & 0.696           \\ \hline
%Vit       & 0.641            & \begin{tabular}[c]{@{}c@{}}BERTC-\\ GCAN-Vit\end{tabular} & 0.704            \\ \hline
%BERTC-Vit & 0.692            & -                                                         & -                \\ \hline
%\end{tabular}
%\caption{Sub-task B soft voting ensemble weighted F1-score for 10-fold cross validation.}
%\label{subtaskbbcv}
%\end{table}

\begin{table}[!htb]
\setlength{\tabcolsep}{4.5pt} % Default value: 6pt
\begin{tabular}{|c|c|c|}
\hline
\textbf{Combination}                    & \textbf{Sub-task A} & \textbf{Sub-task B} \\ \hline\hline
\small{Three uni-modal models} & 0.728      & 0.698      \\ \hline
\small{Four bi-modal models}  & 0.752      & \textbf{0.709}      \\ \hline
\small{All seven models}             & \textbf{0.755}      & 0.706      \\ \hline
\small{Oracle model combination}     & 0.762      & 0.716     \\ \hline
\end{tabular}
\caption{Model-level hard voting ensemble performance with \emph{Setup B} for sub-task A and B.}
\label{subtaskall}
\end{table}

As a last experiment, we applied hard voting on the ensembles.
%is adopted to get the sub-task B results for submission. 
Again, sub-task A results are derived from sub-task B. 

Table~\ref{subtaskall} shows the results of different combinations. 
Generally, the combination of the four bi-modal models in the 2nd row outperforms a combination of three uni-modal models in the 1st row. 
If we combine all uni- and bi-modal models (3rd row), the F1-score is 0.755 for sub-task A, and 0.706 for sub-task B.

The results in bold print represent our submitted approaches for both sub-tasks, showing an F1-score of 0.755 for sub-task A and 0.709 for sub-task B. 

After the challenge ended, we again evaluated all possible subset
combinations of the seven candidate models. The followed combinations give the best achievable results by knowing the official test set reference labels: 
%As an oracle result, based on the official test set reference labels, the combination of 
ViT, BERTC-GCAN-ViT, BERTC-ViT, GCAN-ViT achieves an F1-score of 0.762 for sub-task A, while an ensemble consisting of BERTC-ViT and BERTC-GCAN-ViT yields an F1-score 0.716 on sub-task B. These results are shown for comparison in the final row of Table~\ref{subtaskall} as oracle results.

\section{Conclusion}\label{conclusion}
This paper presents our ensemble-based approach to address two sub-tasks of the SemEval-2022 MAMI competition. The challenge aims to identify misogynous memes and classify them into---potentially overlapping---categories. We train different text models, an image model, and via our proposed fusion network, we combine these in a number of different bi-modal models. 

Among the uni-modal systems, all text models show a far better performance than the image model. As expected, our proposed graph convolutional attention network (GCAN), which also considers the graph structure of the input data while 
using pre-trained RoBERTa word embeddings as node features, consistently outperforms the pre-trained RoBERTa model.

The proposed fusion network further improves the performance by combining the ideas of stream-weighting and representation fusion. We additionally adopt 10-fold cross-validation and use a dataset-level soft voting ensemble to obtain better and more robust results. Finally, our model-level hard voting ensemble integrates the soft voting ensemble predictions of our best uni- and bi-modal models. Our experiments indicate  that this layered ensemble approach can significantly improve the model accuracy. Ultimately, our submitted system results in an F1-score of 0.755 for sub-task A and 0.709 for sub-task B.
%The best ensemble results in an F1-score of 0.762 for sub-task A and 0.716 for sub-task B. 

 Overall, we believe that the identification of misogyny in memes is best addressed through bi-modal recognition, considering both textual and image information. Concerning the text-based classification, we found a graph convolutional attention neural network to be beneficial as an integrative model for Transformer embeddings. This helps in the text classification, when the documents are short, as for the given meme classification task.
 
To cope with the bi-modality of the task at hand, we have implemented a range of systems for integrating the information from both streams. An idea that proved to be effective here was that of bringing together the strengths of early fusion and decision fusion in a joint framework. This allowed us to dynamically adjust the contributions of the two modalities through dynamic stream weighting, while still being able to combine information at the feature level across the streams, thanks to the representation fusion branch of our bi-modal systems.

\newpage

\section*{Acknowledgements}
The work was supported by the PhD School ”SecHuman - Security for Humans in Cyberspace”
by the federal state of NRW.

\bibliography{anthology,custom}

%\appendix
%\input{latex/Sections/Appendix}

%\todo[inline]{We have max. 8 Seiten limitation. So I put the significant plot in Appendix. Acknowledgments, references, and appendices do NOT count toward page limits.}
\end{document}